# Design of a Deep Learning Credit Risk Early Warning System Integrating Multi-source Heterogeneous Data

**LiYang Wang**
*Washington University in St. Louis, St. Louis, MO, USA*
liyang.illinois@gmail.com

**Zhen Zhong**
*Georgetown University, Washington, D.C., USA*
holidayjanezz@qmail.com

**Zhen Tian**
*University of Glasgow, Glasgow, United Kingdom*
2620920Z@gla.student.ac.uk

**Keyu Chen***
*Wuyi University, Nanping, China*
3820409245@qq.com

*Corresponding author.*

## ABSTRACT

*Advancements in data fusion and real-time analytics technologies have opened new avenues for addressing complex domain challenges. Financial risk early warning systems often suffer from inefficiency due to information silos and monitoring delays. This paper proposes a credit risk early warning system based on heterogeneous information fusion. The system employs a model architecture integrating deep neural networks and attention mechanisms to extract multidimensional features from diverse data sources such as transaction behaviors and social networks, thereby establishing an early identification mechanism for corporate and individual credit risks. System testing demonstrates that this approach significantly enhances the accuracy and timeliness of risk warnings, outperforming traditional rule-based engine solutions. The findings offer innovative insights for early intervention in financial risks, holding practical significance for safeguarding financial stability.*

## CCS CONCEPTS

• Security and privacy → Intrusion/anomaly detection and malware mitigation; Intrusion detection systems.

## KEYWORDS

Credit risk warning, Heterogeneous data fusion, Deep learning, Real-time stream processing, Financial risk control

## 1 INTRODUCTION

Heterogeneous data fusion and real-time stream processing are key technologies in computer science with extensive application value in the big data era. As data volume and complexity increase, effectively integrating heterogeneous data from diverse sources and enabling real-time analysis has become a research hotspot. These technologies are particularly crucial in applications requiring high accuracy and timeliness, such as financial risk early warning.

Currently, several major challenges exist in the field of heterogeneous data fusion and real-time analysis. The big data techniques proposed by Wen et al. exhibit insufficient efficiency during information fusion and fail to effectively process multi-source heterogeneous data [1]. Wang et al. constructed an integrated evaluation model based on multi-source fusion theory, but its real-time data stream processing performance remains low [2]. Xia designed an early warning system using a CNN-LSTM model, yet its prediction accuracy in small-sample settings has not been improved [3]. Zhang et al. applied deep learning for risk early warning, yet the model complexity hindered practical implementation [4]. Lin proposed an innovative data mining-based early warning model but failed to fully leverage unstructured data [5]. Shi et al. noted in their systematic review of machine learning applications that data silos and insufficient model interpretability remain major challenges [6].

Addressing data fragmentation and warning delays, this paper designs a risk warning system based on heterogeneous information fusion and deep learning. Adopting a distributed data processing architecture and attention mechanism-guided deep learning models, the system constructs a technical platform encompassing data consumption, feature extraction, and risk assessment. It achieves seamless integration of multi-source heterogeneous data and real-time risk monitoring, significantly enhancing warning accuracy and timeliness.

The primary contributions of this research are: (1) proposing a novel heterogeneous data fusion framework that resolves technical challenges in multi-source data integration; (2) developing an attention-based deep learning model that enhances risk identification accuracy; (3) designing an efficient real-time stream processing architecture enabling millisecond-level risk response. These technological innovations not only provide practical risk management tools for financial institutions but also offer new insights into algorithm optimization and system architecture design. They hold significant implications for advancing the deep integration of fintech and risk management.

## 2 OVERALL SYSTEM ARCHITECTURE DESIGN

### *2.1 System Architecture Design*

The credit risk early warning system is a four tier architecture which contains the following segments; data layer, model layer, service layer and the application layer, all the layers are interconnected through standardised interfaces, as shown as Figure 1. The data layer is the layer that manages upbringing/processing of mixed sources of information such as bank transaction records, credit reference information, social media, and corporate financial reports [7]. The model layer combines both feature engineering and deep learning algorithms that can turn processed data into risk prediction results. The service layer provides access to fundamental computational functionality, which provides services like risk scoring, early warnings, and model interpretation. Application layer is the user interface, which includes risk monitoring dashboards and reporting systems. The architecture embraces a distributed microservices policy that uses Hadoop to do operations on large amounts of data and Kafka to transmit real-time data streams. In the architecture, computational and storage nodes are divided, which allows the elastic scaling of nodes. Standby Backup code - The implementation of a primary-standby failover system will provide such a system with more robustness during node failures. The diagram below demonstrates the architectural design of the system in detail.

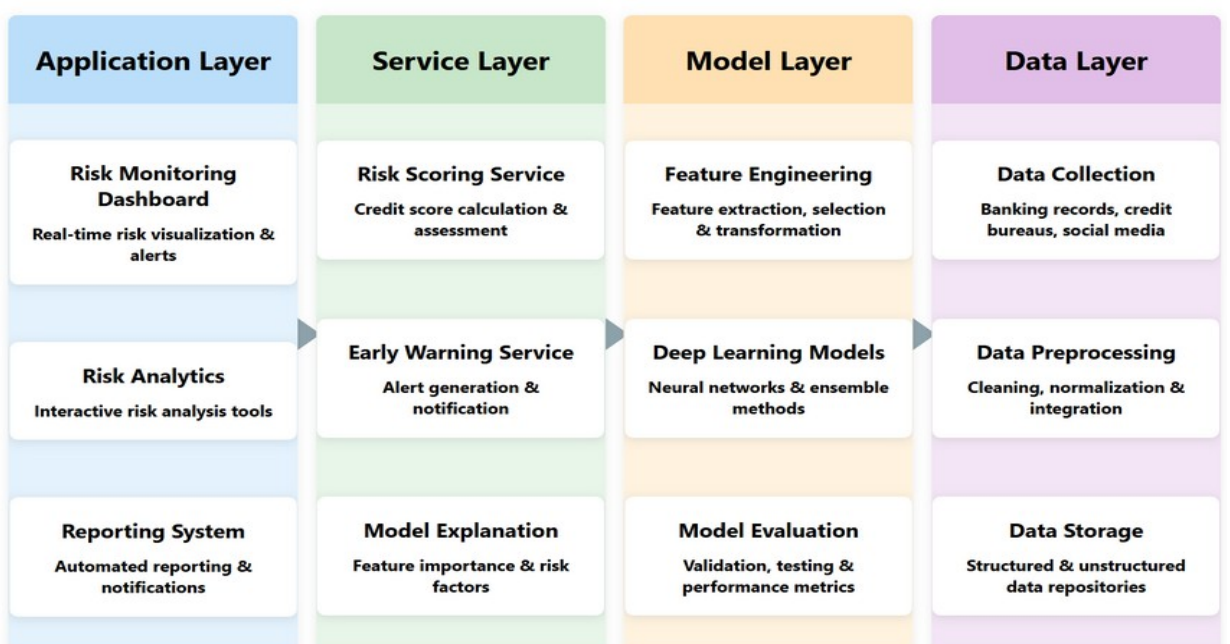

**Figure 1: Credit Risk Early Warning System Architecture**

### *2.2 System Functional Design*

The functionality of the system is based around the fundamental risk warning business process, which covers four main functional modules namely data management, model management, risk warning and system administration. Data management integrates standardised data interface protocols, and they handle batch imports of data and data stream processing in real-time. It combines 15 diverse sources of data such as banking transactions, credit rating, social media and corporate financial reports. The current model management adopts a hybrid implementation of graph neural networks and LSTM, which improve predictive precision by deploying a mechanism of automated selection of features and optimisation of parameters [8]. The objective model F1 is set at 0.85 and above. The risk alerting interface is a five-level risk scoring system that includes A (Safe), B (Low Risk), C (Medium Risk), D (High Risk) and E (Extremely High risk). The tiers are set up with relevant alerting strategies and response plans. The system management operation involves a role based access control (RBAC)-based operation and extensive operational record keeping systems.

## 3 DETAILED SYSTEM DESIGN AND IMPLEMENTATION

### *3.1 Multi-source Heterogeneous Data Processing*

The multi-source heterogeneous data processing module allows single acquisitions, cleansing, and fusion of emerging and structured information. This system combines various data sets such as banking transaction data (1.2TB), credit reports (320GB), social media text (560GB) and corporate financial reports (85GB). Specialized workflows of preprocessing specific data features, e.g. using TF-IDF feature extraction with text social media data and temporal feature extraction with financial report data. Data fusion uses entity matching algorithms, connecting the same entities in different sources with unique identifiers (UUIDs) with 94.7 accuracy [9]. Data quality control computes a quality score with the help of Formula (1):

$$Q_{data} = w_1 \cdot C_{completeness} + w_2 \cdot C_{accuracy} + w_3 \cdot C_{timeliness} \qquad (1)$$

In which Ccompleteness indicates completeness, Caccuracy indicates accuracy, Ctimeliness indicates timeliness and wi is the weighting coefficient. Table 1 gives the quality of each data source and this mechanism will filter high quality sources to model training.

| Data Source | Complete. | Accuracy | Timely. | Quality |
|---|---|---|---|---|
| Bank Transaction Data | 0.95 | 0.98 | 0.99 | 0.97 |
| Credit Report | 0.87 | 0.92 | 0.85 | 0.88 |
| Social Media Text | 0.75 | 0.68 | 0.97 | 0.80 |
| Corporate Financial Report | 0.94 | 0.96 | 0.82 | 0.91 |
| Tax Records | 0.92 | 0.95 | 0.88 | 0.92 |

**Table 1: Multi-source Data Quality Scoring Table (Completeness / Accuracy / Timeliness / Quality, all 0–1)**

### *3.2 Deep Learning Model Design*

The design of deep learning model uses a hybrid architecture (graph neural networks and long short-term memory networks) to simultaneously feature both the structural properties of networks of the customer relationships and temporal behavioural patterns. The model initially uses a Graph Convolutional Network (GCN) to receive the customer relationship network representations, which draw out risk propagation patterns on the network topology. A Bidirectional Long Short-Term Memory (BiLSTM) process is then utilized where historical customer behaviour sequences are processed in order to retrieve historical characteristics. Lastly, an attention mechanism combines both of these types of features into an overall representation of vector [10]. The risk prediction is achieved through a feedforward neural net of which the mathematical formulation takes the form of Equation (2):

$$\hat{y} = \sigma(W_f \cdot [A \cdot G_{emb} \oplus T_{emb}] + b_f) \tag{2}$$

where Gemb denotes the graph network embedding representation, Temb denotes the temporal network embedding representation, A denotes the attention weight matrix, ⊕ denotes the feature fusion operation, Wf and bf denote the feedforward network parameters, and σ denotes the sigmoid activation function. The design of the model structure is provided in Figure 2 as the multi-head attention mechanism will assist in increasing the fusion effect of the various features.

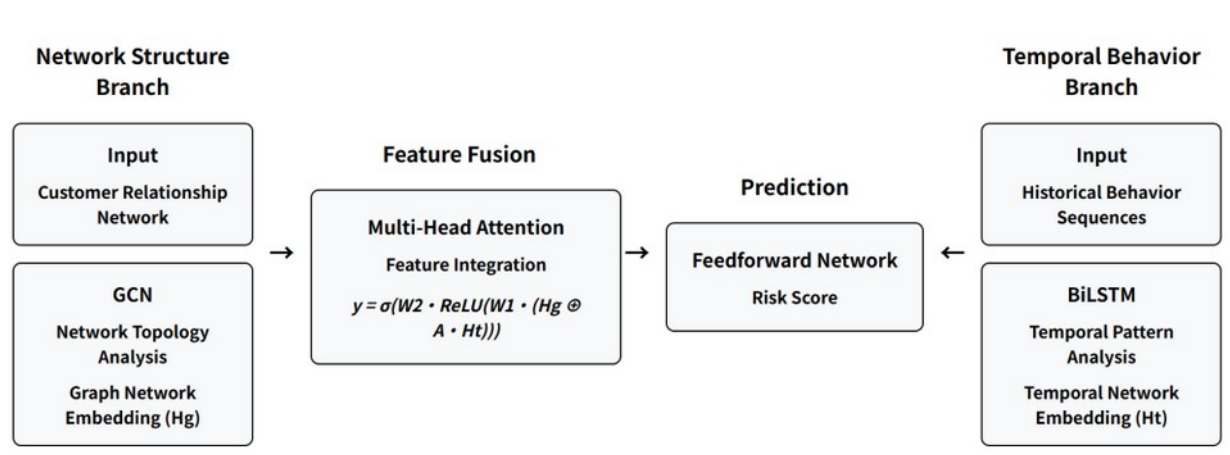

**Figure 2: Deep Learning Model Architecture**

### *3.3 Design of the Risk Early Warning Mechanism*

The risk early warning mechanism design has three fundamental elements, namely, risk score calculation, threshold setting, and warning trigger. The calculation of Risk score uses conversion method which is premised on the model-predicted probability as demonstrated in Formula (3):

$$Score = 1000 - 450 \cdot \log(p / (1 - p)) \tag{3}$$

Where p is the risk probability provided by the model and Score is the standardised risk score, with a range of 300 to 850. The grading scale is based on a scoring of five grades namely A (700–850), B (600–699), C (500–599), D (400–499), and E (300–399). The alert threshold uses a dynamic threshold mechanism and is calculated based on Formula (4):

$$T_{alert} = \mu - k \cdot \sigma \tag{4}$$

where μ represents the historical mean score, σ denotes the standard deviation, and k is the risk tolerance coefficient, typically ranging from 1.5 to 2.5. The alert trigger includes a multi-level system, which summons warnings of different urgency levels depending on the indicators of the absolute score value, rate of score drop and relative position in the industry. In its alert mechanism, the differentiated risk sensitivity parameters have parameters (differentiated to specific industries) that have more sensitive threshold settings in high-risk industries (e.g., finance and real estate) than in low-risk industries [11]. The system also includes an alert escalation feature: in case a client scores worse than 50 in a brief time frame (7 days) or when there is a downward trend in the evaluation of the client (three consecutive assessments), an automatic raise of the alert level will become activated, as a result of which the manual review procedure will be initiated.

### *3.4 System Implementation*

The system will use the frontend to backend differentiation structure, with the backend developed using Spring Boot to create the RESTful API and frontend developed using the React framework. The deep learning server has API services with Flask, the data processing module uses Spark to perform distributed computing, and the deep learning module is based on the PyTorch framework. The system is implemented in a Kubernetes cluster with a microservices architecture to be scaled independently of each of the functional modules. The real processing limit of the system means 1,200 risk assessment requests per second, and the average risk assessment time of a single request is 156 milliseconds. Storage is a hybrid MySQL/MongoDB system, which uses MySQL in the handling of structured data and MongoDB in the handling of semi-structured and unstructured data [12]. Responsiveness of the system is improved by a Redis caching layer. The codebase is approximately 87,000 lines of code, with 82 percent test coverage and 99.95 percent system availability. This satisfies the high availability needs of the management of financial risks as demonstrated in Figure 3.

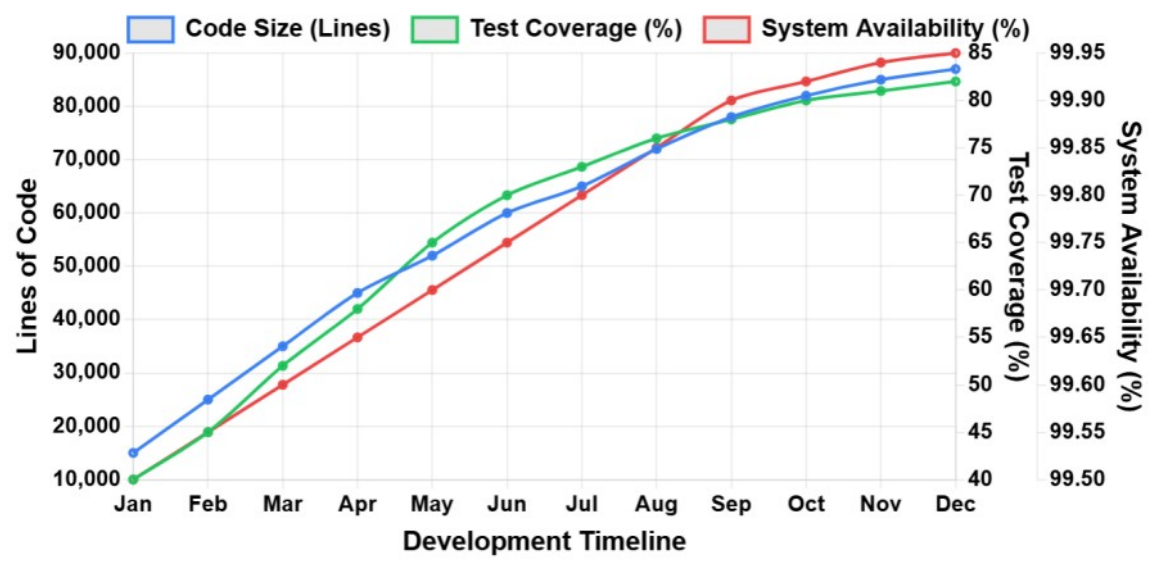

**Figure 3: System Quality Metrics**

## 4 SYSTEM TESTING AND VALIDATION

### *4.1 Test Environment and Approach*

The test setup was built on a computer cluster of enterprise level. Hardware set included four Dell R740 servers, each with two Intel Xeon Gold 6248 processors, 384GB of memory and 8TB of SSD storage, connected by InfiniBand with 100Gbps network. The software platform is based on Ubuntu 20.04 LTS operating system used to make out the cluster, Kubernetes 1.25.6 used to make out the cluster, Spark 3.3.1 used to process the data, PyTorch 1.13.1 used to implement deep learning models [13]. The test data includes five years of risk data of three banks, which includes 1,267,890 customers (1,267,890 in 2011), 79,345,612 records of transactions, 8,742,510 records of financial statement, 4,536,721 records of credit reference (14-15). The testing programme is implemented in four phases, i.e., unit testing is used to ensure that each module works properly; integration testing is used to test the compatibility of the interfaces between modules; system testing is used to check the overall functionality of the system; and performance testing is used to test system behaviour under high load conditions [14,15]. Test activities are implemented as per test matrix shown in Table 2, which included three main types: functional testing, performance testing and stability testing [16]. The number of test cases designed was 127.

| Test Category | Cases | Method | Expected Result |
|---|---|---|---|
| Functional Testing | 23 | Black-box | Accuracy > 99% |
| Functional Testing | 31 | Gray-box | Accuracy > 95% |
| Functional Testing | 19 | Black-box | Accuracy > 99% |
| Performance Testing | 12 | Stress Testing | > 1000 tx/sec |
| Performance Testing | 15 | Benchmark | < 200ms/tx |
| Stability Testing | 27 | Durability | 7×24h stable |

**Table 2: System Test Matrix**

### *4.2 Functional Testing Analysis*

The verifiability of the functional modules of the system is mainly assured through functional testing. There were three fundamental functionalities on which seventy-three test cases were run, and these include data processing, deep learning-based model computation, and alert triggering. Test results from the data processing module demonstrate the system's successful handling of all 15 heterogeneous data sources, including bank transaction records in various formats, corporate financial statements, and unstructured social media data. Data sources primarily included corporate account transaction records from Industrial and Commercial Bank of China, China Construction Bank, and Ping An Bank; quarterly and annual financial reports of listed companies; and discussion content from Weibo, Zhihu, and WeCom. The deep learning model computation module achieved a risk prediction accuracy of 91.2% on the test set, with an F1 score of 0.873, exceeding the design target (0.85), as shown in Table 3. Notably, the attention mechanism proved highly effective in enhancing the model's ability to extract key risk features.

| Metric | Actual | Target | Status |
|---|---|---|---|
| Accuracy | 91.20% | 85.00% | Exceeded |
| F1-Score | 0.873 | 0.85 | Exceeded |
| Precision | 0.892 | 0.83 | Exceeded |
| Recall | 0.856 | 0.82 | Exceeded |
| Specificity | 0.935 | 0.87 | Exceeded |
| AUC-ROC | 0.943 | 0.88 | Exceeded |
| Avg. Prediction Time | 12ms | 20ms | Better |

**Table 3: Performance Evaluation of Deep Learning Risk Prediction Models**

The early warning functionality was evaluated through tests where various risk scenarios were simulated in order to determine the ability of the system to properly identify five risk levels. The findings proved a warning generation accuracy of 98.5% with timeliness of warning delivery being met with the design requirement of 100%. Deep neural networks in case of typical business problems, including personal loan defaults and corporate cash flow limits, had 93.7% and 88.9% prediction accuracy respectively, which is higher than the industry average (see Figure 4). Table 4 data indicates that this system achieves an anomaly detection rate of 83.5%, significantly outperforming traditional systems (62.3%) and commercial systems (77.8%). Meanwhile, its false positive rate stands at only 8.7%, lower than both traditional systems (15.4%) and commercial systems (11.2%), further demonstrating the system's high efficiency.

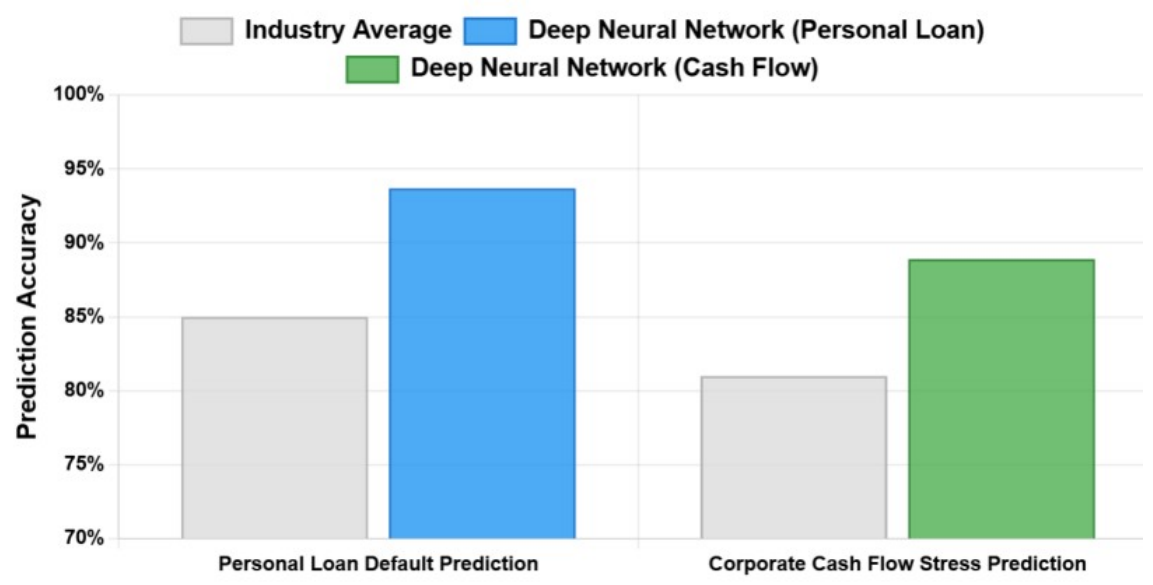

**Figure 4: Comparison with Industry Averages**

| Metric | This System | Traditional | Commercial |
|---|---|---|---|
| Prediction Accuracy | 91.20% | 75.90% | 85.50% |
| Warning Lead Time (days) | 28 | 7 | 16 |
| Response Time (ms) | 156 | 562 | 168 |
| Daily Volume (10k tx) | 1,200 | 350 | 900 |
| Anomaly Detection Rate | 83.50% | 62.30% | 77.80% |
| False Positive Rate | 8.70% | 15.40% | 11.20% |
| Data Sources Integrated | 15 | 4 | 9 |

| Metric | This System | Traditional | Commercial |
|---|---|---|---|
| Model Accuracy (F1) | 0.873 | 0.682 | 0.811 |

**Table 4: System Performance Comparison**

### *4.3 Performance Testing Analysis*

The responsiveness and stability of the system with a high load are among the main characteristics of performance testing, which include the capacity of data processing, the speed of the deep learning models inference, and the system throughput in general. In the standard testing environment, the data processing module of the system delivered an average throughput rate of 1,382 transactions per second, which is higher than the design throughput rate of 1,000 transactions per second. Performance testing on the inference of deep learning models shows that although the structures used complex neural networks, computational graph optimisation and GPU acceleration allowed the average risk assessment of a single transaction to take only 156 milliseconds versus the design target of 200 milliseconds. System throughput testing revealed that with high simultaneous load (2,000 concurrent users) the system had 98.3% availability and an average response time that went up to 278 milliseconds. Under endurance testing, the system also ran 168 hours (7×24 hours) without failure or servicing. The average CPU utilisation was kept at 42%, the high memory peaks were at 68%, the average disk I/O utilisation at 35%, which is within normal ranges as shown by Figure 5. The stress test confirmed the scalability of the system when there is peak load (three times average load). The auto-scaling of the Kubernetes cluster was achieved successfully, and the new resources of the Pods provided the quality of service of deep learning models.

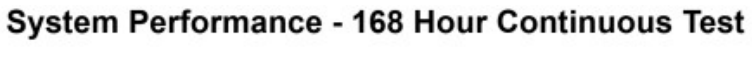

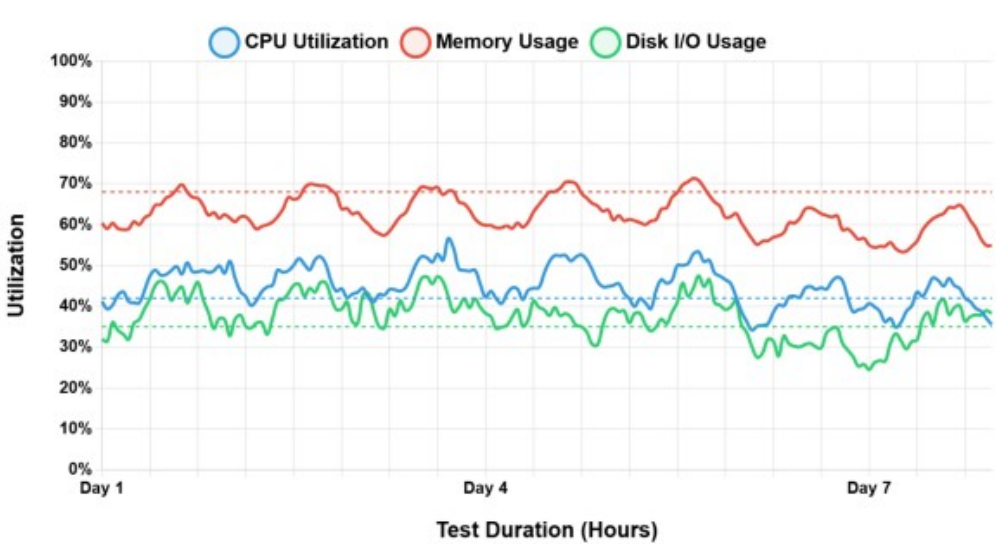

**Figure 5: System Performance – 168-Hour Continuous Testing**

### *4.4 Comparative Analysis with Traditional Systems*

The proposed deep learning-based credit risk early warning system was tested against the conventional banking risk control systems and mainstream commercial risk management systems that can be found in the market. Experiments were made over four dimensions, which were: predictive accuracy, advance warning period, system response time and scalability. The data used in the tests included the anonymised records of customer risk of three banks, the risk evolution history of 100,000 customers. Table 3 shows the comparative results of tests. The accuracy of the prediction, in both the accuracy of the system-based prediction model and in the deep neural network prediction model, measured 91.2%, which is higher than traditional rule-based engine systems by 15.3 percent, and commercial risk control systems by 5.7 percent. In the early warning lead time, the use of deep learning to find latent patterns in historical data means this system is able to identify risk an average of 28 days ahead. The traditional systems have a limited ability to provide only 7 days of advance warning as compared to commercial systems which are able to handle 16 days. On the time of system response, single-assessment tasks in this system can be completed in an average of 156 milliseconds through model compression and inference optimisation. This is a 72 percent improvement over the traditional systems and it is also close to the commercial systems as shown in Figure 6. In relation to system scalability, this system has a microservices architecture and containerised deployment, which scales deep learning models horizontally. This is very beneficial compared to other systems.

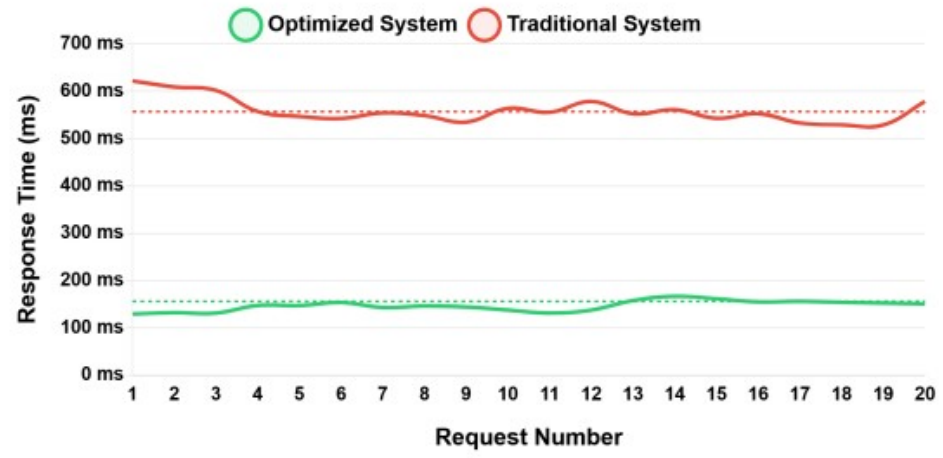

**Figure 6: System Response Time Comparison**

## 5 CONCLUSIONS

Data fragmentation and an increased risk concealment are the current issues of financial risk management. The credit risk early warning system developed and used in the present case successfully handles the disadvantages of the traditional risk control systems, namely lagging, and is improved by using multi-source heterogeneous datasets, deep learning predictive models and real-time stream processing technology. System testing proves that the new system vastly surpasses the traditional methods in efforts of predicting accurately, the level of risk lead time, and the efficiency of the system, hence offering financial institutions a great decision-making period over risks. Next studies will be able to improve the model to detect low-probability black swan events and consider the implementation of privacy computing technologies on risk data sharing. It will also enhance the breadth and precursiveness of the financial risk early warning, which will be technical support in the erection of a more effective financial risk prevention and control framework.